\documentclass[a4paper,english]{lipics-v2018}
\usepackage{microtype}
\usepackage{subfigure}
\title{Modelling Irregular Spatial Patterns using Graph Convolutional Neural Networks}

\titlerunning{Modelling Spatial Patterns using Graph Convolutional Networks}

\author{Di Zhu}{Institute of Remote Sensing and Geographical Information Systems, Peking University, 5th Yiheyuan Road, Beijing, China}{patrick.zhu@pku.edu.cn}{http://orcid.org/0000-0002-3237-6032}{}

\author{Yu Liu}{Institute of Remote Sensing and Geographical Information Systems, Peking University, 5th Yiheyuan Road, Beijing, China}{*Correspondence: liuyu@urban.pku.edu.cn}{http://orcid.org/0000-0002-0016-2902}{}

\authorrunning{D.\ Zhu, Y.\ Liu}

\Copyright{Di Zhu and Yu Liu}

\subjclass{Information systems $\rightarrow$ Geographic information systems}


\category{arXiv preprint}

\funding{This research was supported by the National Key Research and Development Program of China [Grant Number: 2017YFB0503602] and the National Natural Science Foundation of China [Grant Number: 41625003].}

\nolinenumbers 
\hideLIPIcs  

\begin{document}

\maketitle

\begin{abstract}
The understanding of geographical reality is a process of data representation and pattern discovery. Former studies mainly adopted continuous-field models to represent spatial variables and to investigate the underlying spatial continuity/heterogeneity in the regular spatial domain. In this article, we introduce a more generalized model based on graph convolutional neural networks (GCNs) that can capture the complex parameters of spatial patterns underlying graph-structured spatial data, which generally contain both Euclidean spatial information and non-Euclidean feature information. A trainable semi-supervised prediction framework is proposed to model the spatial distribution patterns of intra-urban points of interest(POI) check-ins. This work demonstrates the feasibility of GCNs in complex geographic decision problems and provides a promising tool to analyze irregular spatial data.
 \end{abstract}
\section{Introduction}\label{section:intro}
\begin{figure}[t]
	\centering
	\includegraphics[width=\textwidth]{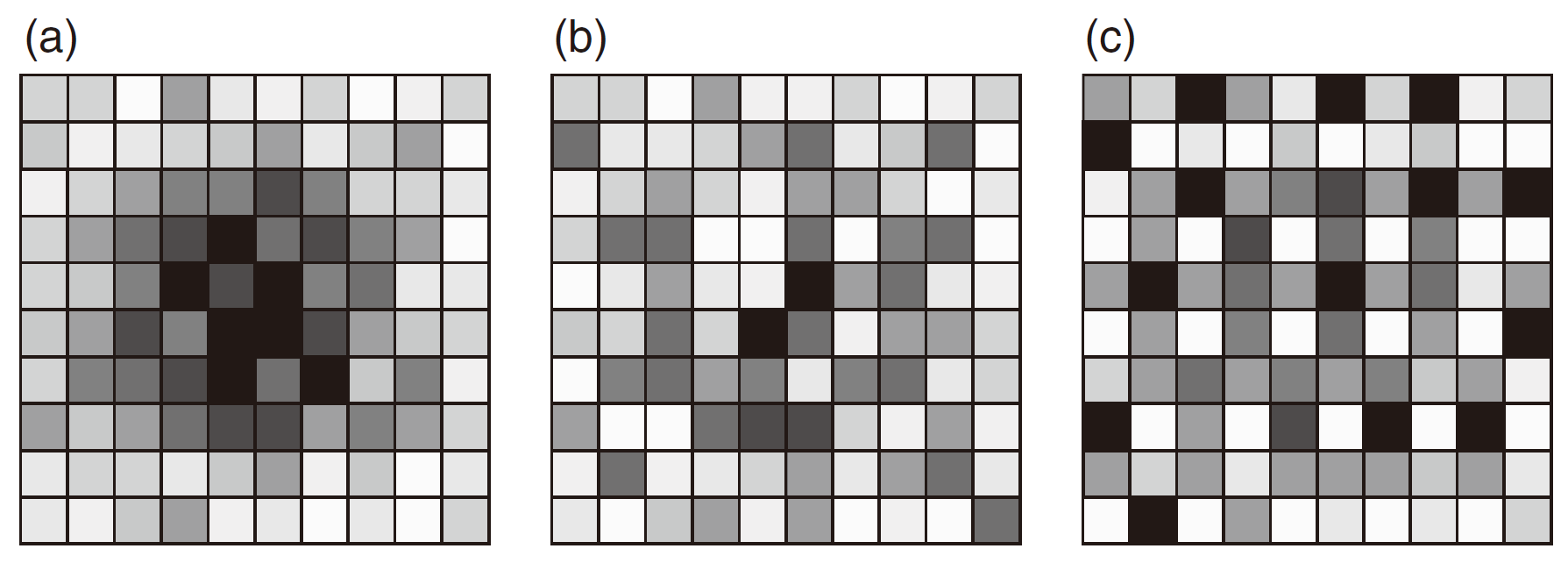}
	\caption{Spatial patterns represented in a regular grid \cite{fotheringham2008sage}. (a) Positive spatial autocorrelation. (b) Spatial randomness. (c) Negative spatial autocorrelation.}
	\label{fig:spatial patterns}
\end{figure}

The continuous-field model, which can be seen as a process of reducing the number of spatial variables required to represent reality to a finite set (a field) \cite{Goodchild1992Geographical}, is a fundamental perspective in modelling the complex geographical world. The variation of attributes in a field model represents the spatial pattern of certain geographical phenomenon at the conceptual level of abstraction \cite{Liu2008Towards, Goodchild2007Towards}, as is shown in Figure \ref{fig:spatial patterns}. The analysis of spatial patterns based on field models has been studied extensively in traditional geography applications \cite{cressie_origins_1990, Ord1995Local}. Methods can be roughly divided into two types: autoregressive methods that adopt a spatial lag term to consider the autocorrelation of local neighborhoods \cite{Anselin1992SPATIAL} and geostatistical methods that use semi-variograms to characterize the spatial heterogeneity \cite{Matheron1963Principles, cressie_origins_1990}.

To uncover the deep features of spatial patterns, convolutional neural networks (CNNs) have been introduced from computer science to investigate local stationary properties of the input data by allowing long range interactions in terms of shorter, localized interactions \cite{Lecun2015Deep}. However, the use of CNNs becomes problematic when the data is not structured in the regular spatial domian (e.g. raster model in GIS), since the local kernel filter can no longer be defined via the Euclidean metric of the grid. 
Graph convolutional networks (GCNs) is a generalization of CNNs to deal with graph-structured data in the irregular spatial domain (i.e., vector model in GIS), where the input data is represented as objects and their connections. The convolutional filter in GCNs can be extended to be localized in the spectral domain of the objects' features \cite{Defferrard2016Convolutional, Henaff2015Deep}, thus enables the investigation of spatial dependence in the irregular spatial domain. We think that GCNs are suitable for modelling the complex spatial patterns in geographical data that generally contain both Euclidean spatial information and non-Euclidean feature information \cite{liu2015social}. 

In this article, we introduce a way to model the spatial patterns in geographical data by constructing graph neural networks with both spatial information and feature information embedded and by designing a localized feature filter on the graph that considers spatial constraints. A layer-wise semi-supervised neural network framework is proposed to make the model trainable. In addition, we have applied the proposed model in an intra-urban prediction case based on a POI check-in dataset in Beijing, China to demonstrate the feasibility of our model.

\section{Embedding spatial patterns in graphs}\label{section:method}
\subsection{Graph Fourier transformation}
To enable the formulation of fundamental operations such as filtering on a graph, the Graph Fourier transform is needed first, which is defined via a generalization of the Laplacian operator on the grid to the graph Laplacian \cite{Fan1997Spectral}. In graph $G=(V,E,W)$, $V$ is a finite set of $|V|=n$ nodes, $E$ is a set of edges among nodes and $W\in \mathbb{R}^{n \times n}$ is a weighted adjacency matrix representing the weights of edges. An input vector $x\in \mathbb{R}^n$ is seen as a signal defined on $G$ with $x_i$ denotes the spectral information of node $i$.
\begin{definition}[Graph Laplacian]
	\label{definition:Laplacian}
	Let $L=\Delta-W$ be the graph Laplacian of $G$, where $\Delta\in \mathbb{R}^{n\times n}$ is a diagonal matrix with $\Delta_{ii}=\sum_j W_{ij}$, and the normalized definition is $L^s=I_n-\Delta^{-1/2}W\Delta^{-1/2}$ where $I_n$ is the identity matrix.
\end{definition}

As $L^s$ is a real symmetric positive semidefinite matrix, it has a complete set of orthonormal eigenvectors $U=(u_1,\cdots,u_n)$, and their associated nonnegative eigenvalues $\lambda=(\lambda_1,\cdots,\lambda_n)$. The Laplacian is diagonalized by $U$ such that $L^s=U\Lambda U^T$ where $\Lambda=diag([\lambda_1,\cdots,\lambda_n])\in \mathbb{R}^{n\times n}$. The graph Fourier transform of $x\in \mathbb{R}^n$ is then defined as $\hat{x}=U^T x\in \mathbb{R}^n$.

\subsection{Convolutions on graphs}
\begin{definition}[Graph convolutions]
	\label{definition:graphconvolution}
	The convolution operators on graphs are defined as the muliplication of $x$ with a filter $g_\theta=diag(\theta)$ parameterized by $\theta \in \mathbb{R}^n$ in the Fourier domain, i.e.:
	\begin{equation}
	\label{eq:g_theta}
	g_\theta \star x=g_\theta(L^s)x=g_\theta(U\Lambda U^T)x=Ug_\theta(\Lambda)U^Tx.
	\end{equation}
	We can understand $g_\theta(\Lambda)$ as a function of the eigenvalues of $L^s$, a non-parametric filter whose parameters are all free and can be trained.
\end{definition}

However, the evaluation of Eq. \ref{eq:g_theta} is computationally expensive, as the multiplication with eigenvector matrix $U$ is $\mathbb{O}(n^2)$. To overcome this problem, \cite{Hammond2009Wavelets} suggested the Chebyshev polynomials $T_k(x)=2x T_{k-1}(x)-T_{k-2}(x)$ up to $K^{th}$ order to approximate $g_\theta(\Lambda)$:
\begin{equation}
\label{eq:g_theta_approximate}
g_{\theta'}(\Lambda)\approx\sum\limits_{k=0}^{K}{\theta_k'} T_k(\tilde{\Lambda}),
\end{equation}
with a rescaled $\tilde{\Lambda}=\frac{2}{\lambda_{max}}\Lambda-I_n$, $\theta' \in \mathbb{R}^K$ is a vector of polynomial coefficients, $T_0(x)=1$ and $T_1(x)=x$.

Furthermore, by assuming $K=1$ and $\lambda_{max}=2$ in Eq. \ref{eq:g_theta_approximate} and some renormalization tricks, \cite{Kipf2016Semi} proposed an expression with a single parameter $\theta=\theta_0'=-\theta_1'$ to compute:
\begin{equation}
\label{eq:g_theta_kipf}
g_\theta \star x \approx \theta(I_n+\Delta^{-1/2}W\Delta^{-1/2})x=\theta \tilde{\Delta}^{-1/2}\tilde{W} \tilde{\Delta}^{-1/2} x,
\end{equation}
where $\tilde{W}=W+I_n$ and $\tilde{\Delta}_{ii}=\sum_j\tilde{W}_{ij}$. Eq. \ref{eq:g_theta_kipf} has complexity $\mathbb{O}(|E|)$ because $\tilde{W}x$ can be efficiently implemented as a product of a sparse matrix with a dense vector.

\subsection{Spatial-enriched graph construction}
Different from state-of-the-art graph constructions in many recognition tasks, where the adjacency matrix $W$ are often defined by calculating the similarity among nodes, we try to enable the constructed graph to capture the relationships between the feature similarity and the spatial displacement of node pairs, i.e., to construct a spatial-enriched graph.

Given the input features $X\in \mathbb{R}^{N\times C}$ of nodes $V$, where $N=|V|$ is the number of locations and $C\in \mathbb{R}$ is the number of features for each node, we can define the adjacency matrix $W$ according to the spatial displacement of $N$ locations. The distance matrix for locations can be considered a prior knowledge for the graph construction process and we can introduce the distance decay effect in geography to represent the spatial dependence of features in $X$.  Derived from the gravity model, there many functions that could be used to express the spatial weighting function, such as binary adjacency judgement, the power function, the exponential function, and the Gaussian function \cite{DiZhu2018Inferring}. For example, considering a variant of the self-tuning Gaussian diffusion kernel \cite{Henaff2015Deep}:
\begin{equation}
\label{eq:gaussian}
W_{ij}=exp^{-\frac{d(i,j)}{\sigma_i \sigma_j}},
\end{equation}
where $d(i,j)$ is the Euclidean distance between node $i$ and $j$ and $\sigma_i$ is computed as the distance $d(i,i_k)$ corresponding to the $k$-th nearest neighbor $i_k$ of node $i$. Eq. \ref{eq:gaussian} gives a normalized measurement of spatial displacement in a graph whose variance is locally adapted around each location. 

Compared to traditional geographical studies that choose arbitrary models to capture the effect of distance, our GCN-based model is a more universal way to model the relationship underlying spatial data. We treat the feature information and the spatial information separately, and leave the graph to learn the spatial pattern given certain training objective. The details of learned spatial pattern are restored in the layer-wise parameters of the deep graph convolutional network and can be adopted in various applications.

\section{Experiment: prediction of intra-urban check-in patterns}\label{section:experiment}
\subsection{Prediction framework}\label{sec:Framework}

Based on the spatial-enriched graph convolutional methods introduced in Section \ref{section:method} that can learn the heterogeneity pattern underlying irregular spatial data, we design a trainable semi-supervised framework for the prediction of check-ins of intra-urban point of interests (POI). The  framework is an example to show how the graph convolutional model can be adopted in geographic decision problems by trying to capture the underlying properties of spatial patterns.

Formally, the goal of the framework is to learn a complex function of the spatial pattern on a graph $G=(V,E)$, which takes as input:
\begin{itemize}
	\item A feature matrix $X\in \mathbb{R}^{N \times C}$ that contains the features $x_i$ for every observed node $i$, where $N$ is the number of given nodes and $C$ is the number of input feature channels
	\item A fully-connected spatial distance matrix $W\in \mathbb{R}^{N \times N}$ summarized using Eq. \ref{eq:gaussian} that represents the spatial structure of observed nodes
\end{itemize}
and outputs a vector $Z=[Z_1,\cdots,Z_N]\in \mathbb{R}^N$ that contains the predicted value for each nodes. 

\begin{figure}[!htp]
	\centering
	\includegraphics[width=0.8\textwidth]{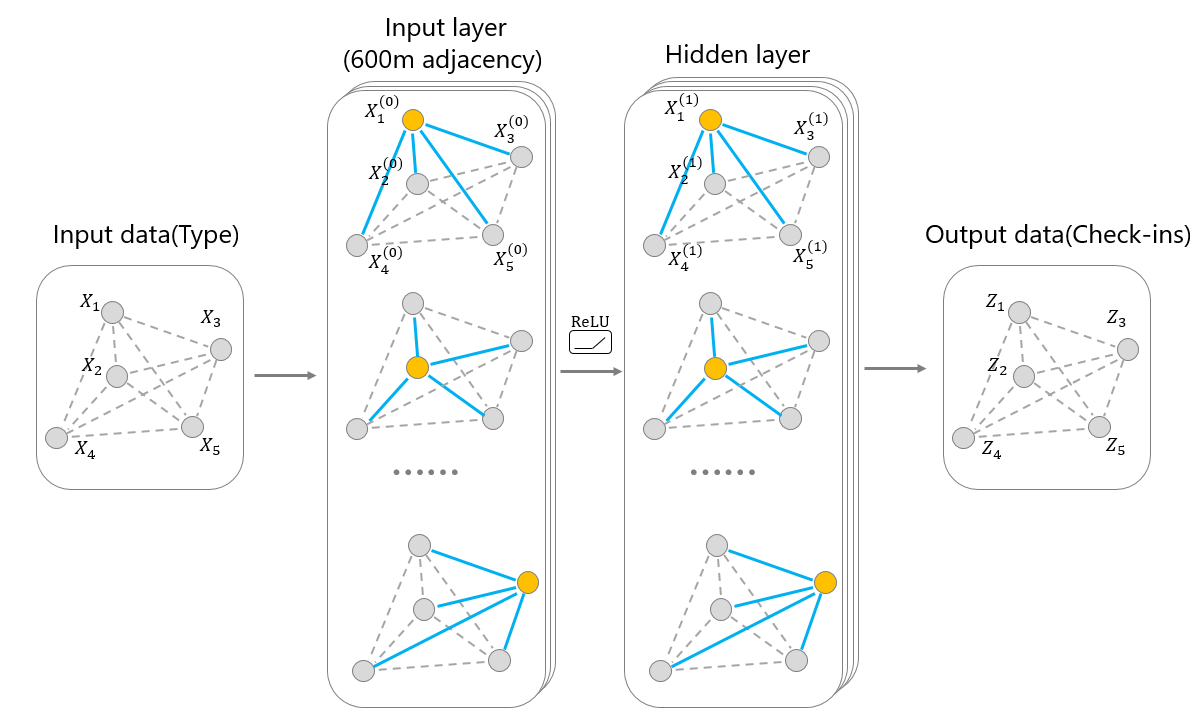}
	\caption{Illustration of the semi-supervised framework for spatial pattern prediction based on graph convolutional networks.}
	\label{fig:GCN framework}
\end{figure}

In the case of the check-in prediction, the graph $G$ is constructed by the spatial displacement of POIs. The input signal is set to be the one-channel feature matrix $X \in \mathbb{R}^N$ that contains the functional types of all POIs. The semi-supervised training means that we train the model according to actual check-in numbers of only a small portion of training POIs and attempt to reproduce the overall spatial pattern of all the POIs, as is shown in Figure \ref{fig:GCN framework}.

For simplicity, we adopt a simple two-layer GCN to capture the spatial dependence among POIs. Recalling the convolutional filter introduced in Eq. \ref{eq:g_theta_kipf}, let $\widehat{W}=\tilde{\Delta}^{-1/2}\tilde{W} \tilde{\Delta}^{-1/2}$, the forward propagation then takes the simple form:
\begin{equation}
Z=\widehat{W} ReLU \left( \widehat{W}X \Theta^{(0)} \right) \Theta^{(1)},
\end{equation}
where $\Theta^{(0)}\in \mathbb{R}^{C\times H}$ is the input-to-hidden parameters for a hidden layer with $H$ feature maps.  $\Theta^{(1)}\in \mathbb{R}^{H\times 1}$ is the hidden-to-output parameters for an output predicted vector $Z$. Note that we do not add any activation function after the second layer because the framework is designed to predict the actual regression numbers of check-ins instead of the discrete classes of POI types.

\subsection{Data descriptions}
\begin{figure}[!htp]
	\subfigure[Spatial pattern of POI types]{\label{fig:data-a}\includegraphics[width=0.5\textwidth]{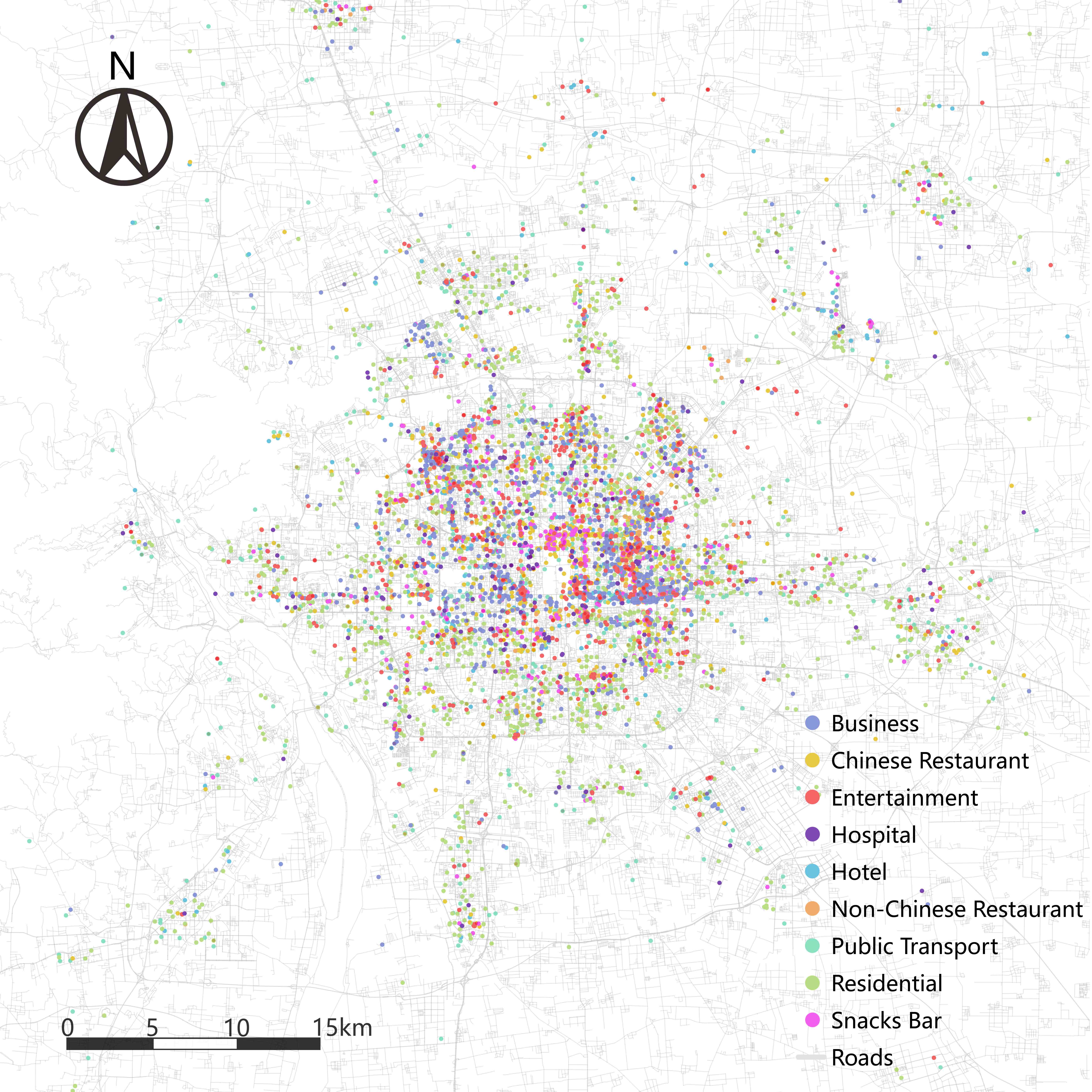}}
	\subfigure[Distribution of the check-in numbers]{\label{fig:data-b}\includegraphics[width=0.5\textwidth]{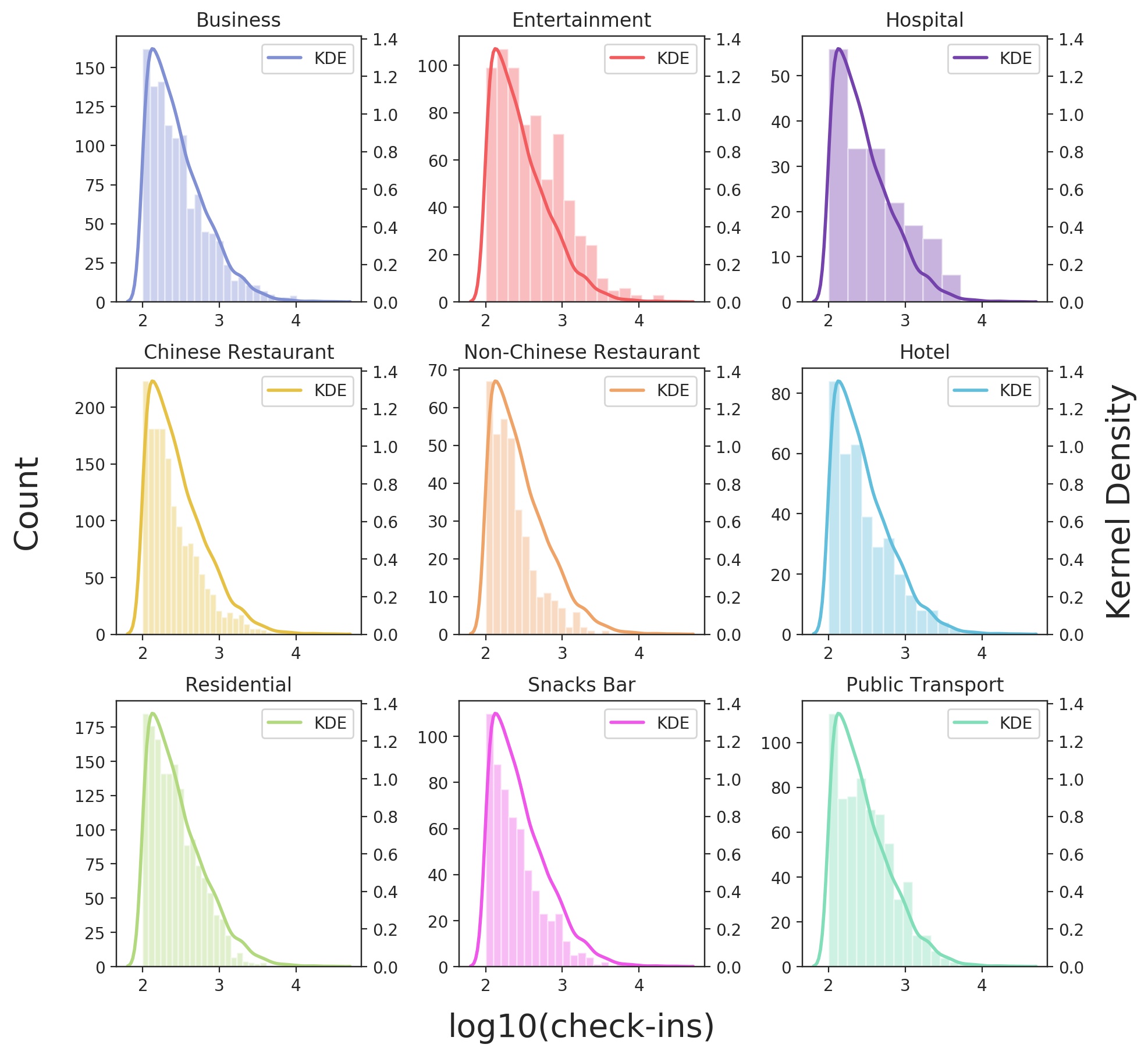}}
	\caption{(a) Spatial pattern of the POIs that contains more than 100 annual check-in records and their functional types. (b) Statistical distributions of check-in numbers for each type in the logarithmic x-coordinate.}
	\label{fig:data}
\end{figure}
We utilized a dataset collected from Sina Weibo in 2014 that contains 868 million check-in records for 143,576  points of interest (POIs) in Beijing \cite{Long2013How}. The original dataset include the POI types (242 in total) and the annual check-ins for each points. After preprocessing, we extracted 6781 typical POIs that have more than 100 check-in records in 2014 and reclassified these POIs into nine functional types according to the 242 subtypes (Figure \ref{fig:data-a}): business, entertainment, hospital, Chinese restaurant, non-Chinese restaurant, hotel, residential, snacks bar and public transport. The spatial distribution of these typical POIs generally covers the urban areas within Beijing. The statistical distribution of the check-in numbers for each types are plotted in Figure \ref{fig:data-b}, where we find a common heavy-tail pattern for all the nine types, indicating that most POIs are not very active and only a few POIs have high check-in numbers.

We constructed the spatial adjacency matrix for the 6781 POIs by generating a 600m buffer for each point. If two POIs are contained in any buffer area, an unweighted edge is added to the graph connecting these two nodes. In this way, the spatial distance matrix defined in Section \ref{sec:Framework} collapses into a binary, unweighted adjacency matrix indicating whether two points are close ($W_{ij}=1$) or not ($W_{ij}=0$). Although this is a simplification for the graph convolution, we find it enough to obtain satisfying result.

\subsection{Semi-supervised training}
For each training, we randomly select 5\% of the POIs as our training set and use the other 95\% POIs as the validation samples. The proposed GCN model in Figure \ref{fig:GCN framework} is fed with the types and the adjacency matrix of all POIs as input. The real check-in numbers of POIs in the training set are utilized to optimize the hyper-parameters through the Adam gradient descent \cite{Kingma2014Adam}. We do not use the validation check-in numbers for training.

We train each graph convolutional model for 2000 epochs, i.e. back propagate and update the parameters for 2000 times with a $3\times 10^{-4}$ learning rate, $5\times 10^{-5}$ L2 regularization and 0.2 dropout rate. We initialize weights using the initialization described in \cite{Glorot2010Understanding} and accordingly normalize the input one-hot vectors of POI types, the hidden layer is set to have 32 hidden units. We choose the L1Loss function to compute the loss for each training epoch, i.e. $L=\sum_i|Z^*_i-Z_i|, i\in \mathbb{T}$ , where $Z^*$ and $Z$ are the actual check-in numbers and predicted check-in numbers, separately (both after log transformation), and $\mathbb{T}$ is the index set of the training POIs.

\begin{figure}[!htp]
	\includegraphics[width=0.6\textwidth]{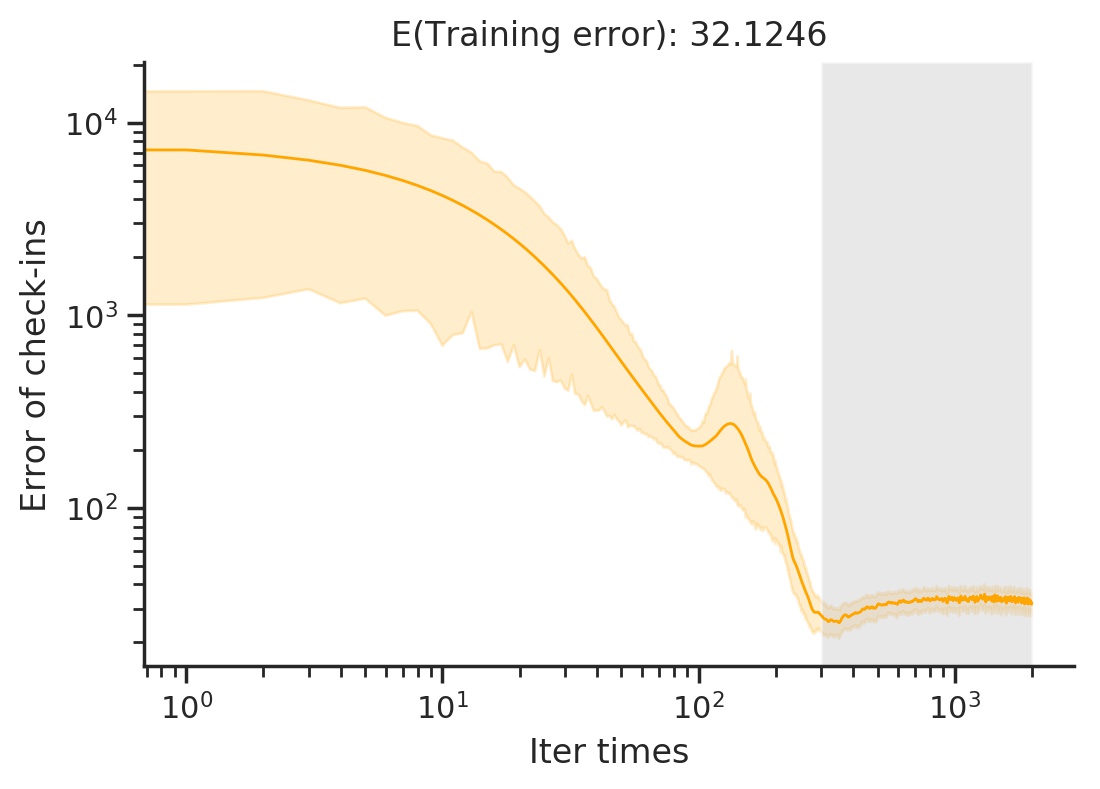}
	\caption{Errorplots for 50 trainings with different initial conditions. Mean error is plotted as the orange curve and the variance of errors is shown by the upper and lower bounds.}
	\label{fig:training}
\end{figure}

We further report the absolute errors of check-in numbers during the training process for 50 random experiments with different initializations and different 5\% training sets (randomly selected), the results are plotted in Figure \ref{fig:training}. Since different training POIs may greatly influence the given pattern of check-ins, training errors appear unstable during the first 200 epochs. While after 200 epochs, the check-in errors remain at a low level. The error finally converges at around 32 by 2000 epochs. The semi-supervised prediction with different initial conditions shows that our GCN model can achieve a high accuracy (less than 10\% error of the average check-in number for a location) no matter the composition of training samples.

\subsection{Predicted spatial patterns}\label{sec:predicted_patterns}
\begin{figure}[!htp]
	\subfigure[]{\label{fig:predicted_all_a}\includegraphics[width=0.35\textwidth]{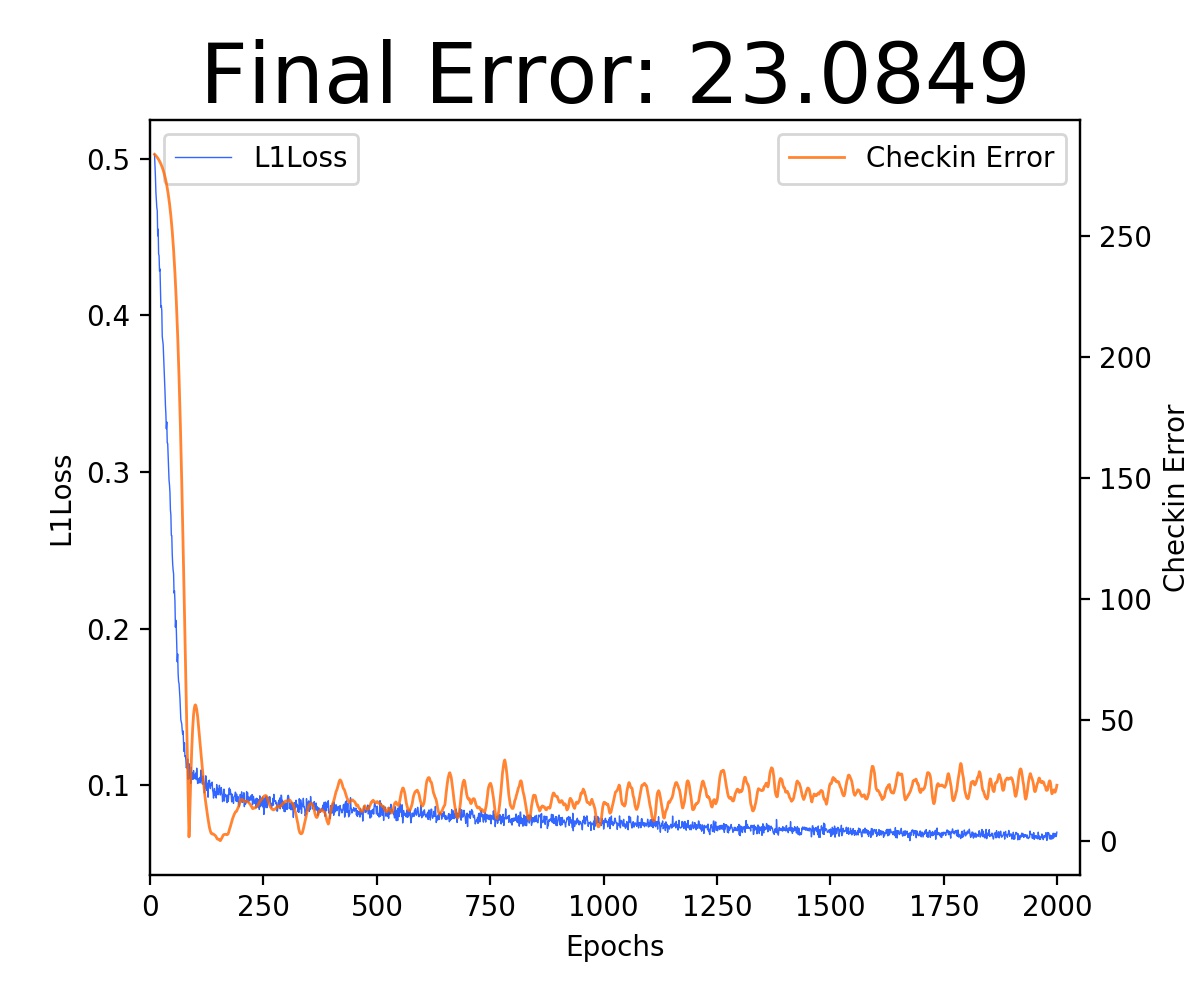}}
	\subfigure[]{\label{fig:predicted_all_b}\includegraphics[width=0.66\textwidth]{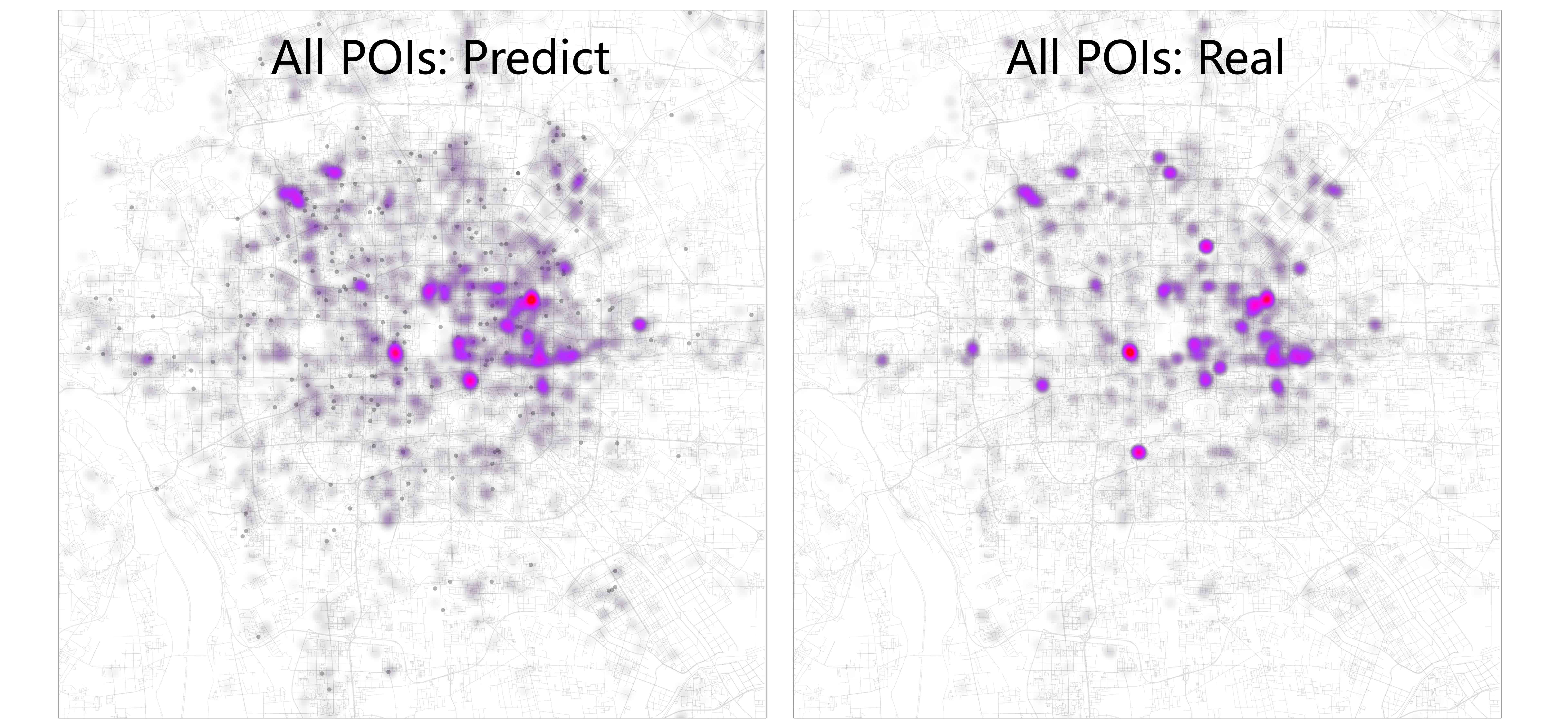}}
	\caption{Results of an example experiment. (a)L1Loss and absolute error during the training procedure. (b)Heatmaps of the predicted check-in pattern and the real check-in pattern.}
	\label{fig:predicted_results_all}
\end{figure}

\begin{figure}[!htp]
	\centering
	\includegraphics[width=0.75\textwidth]{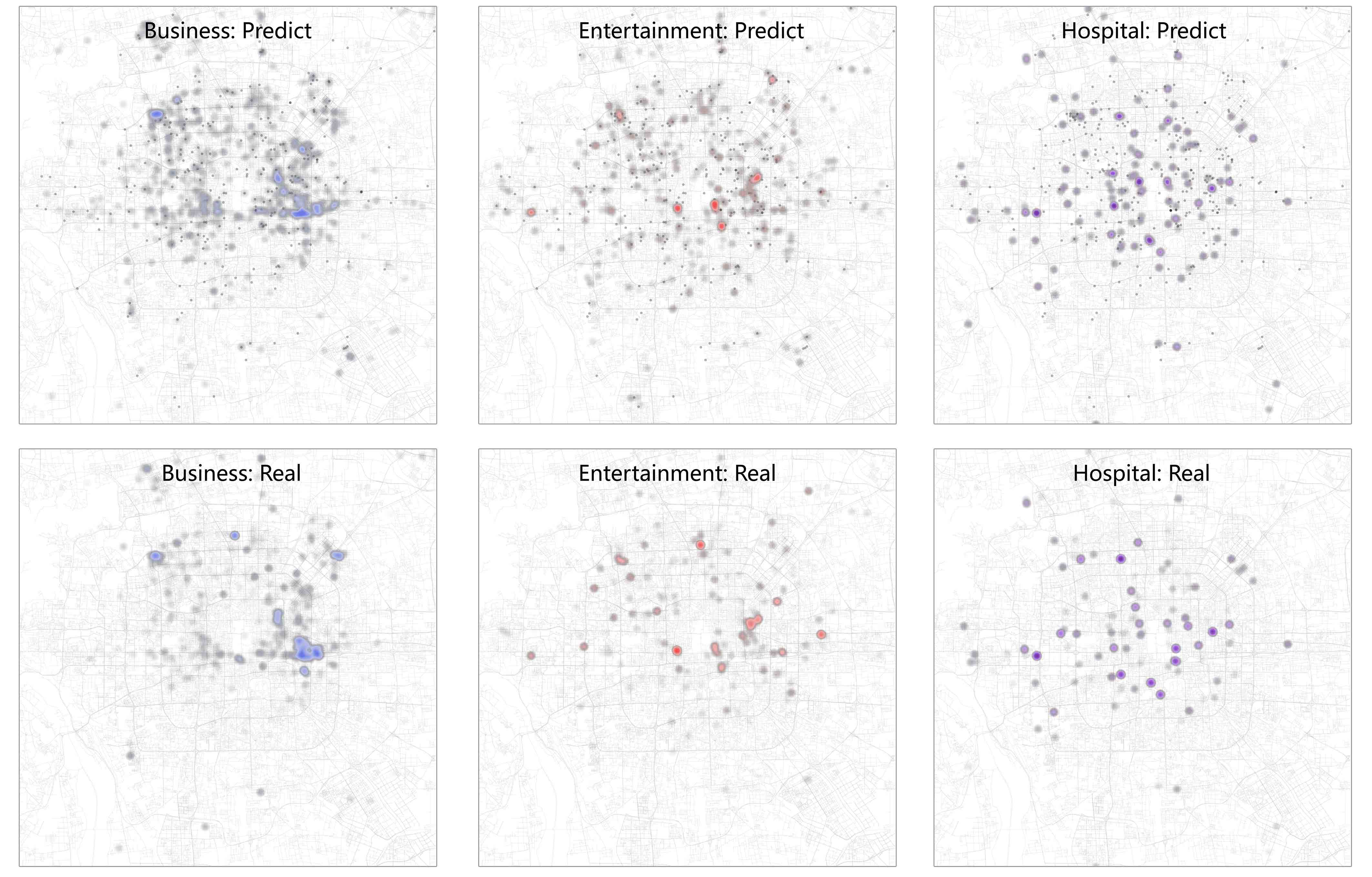}
	\includegraphics[width=0.75\textwidth]{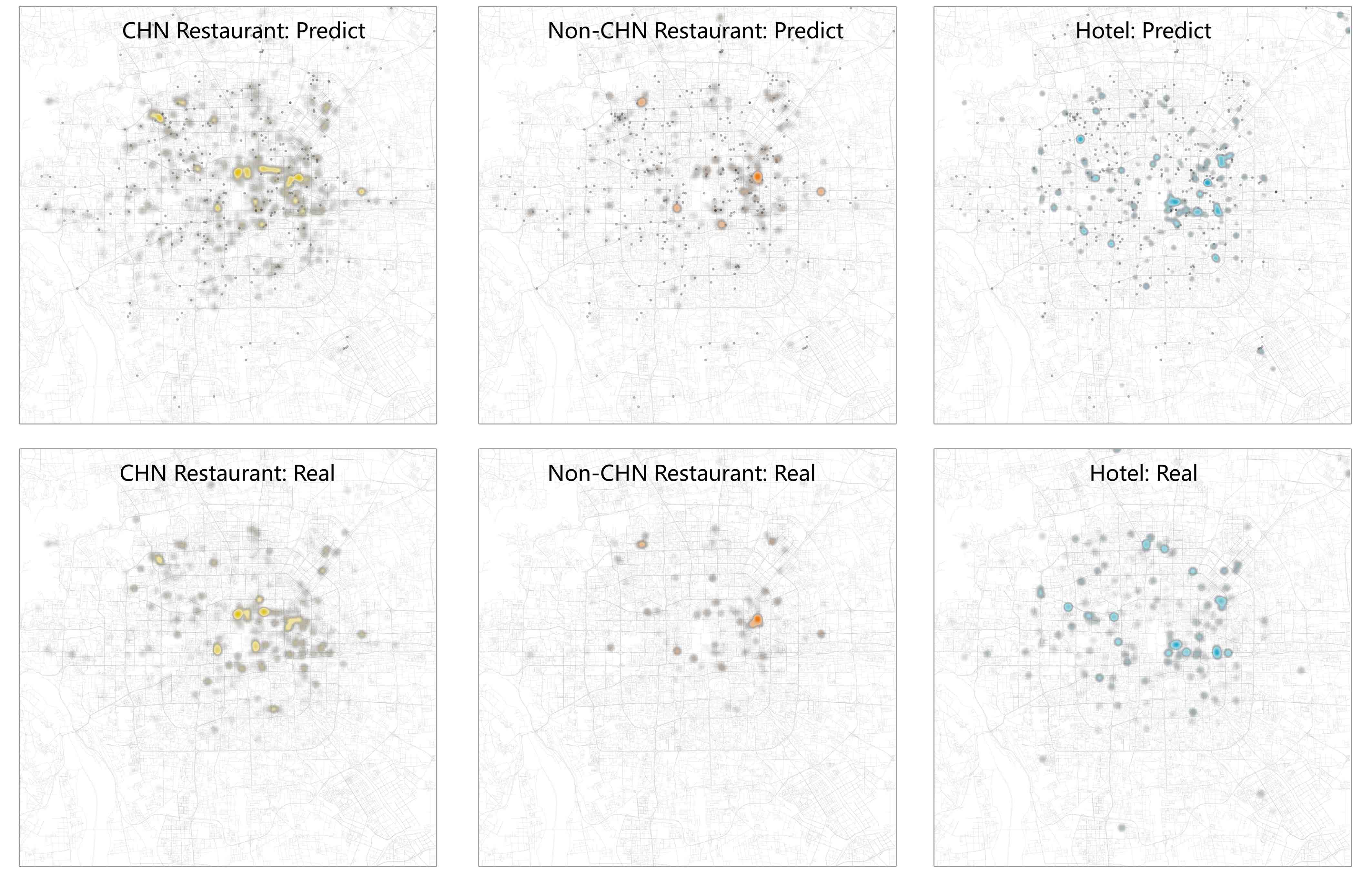}
	\includegraphics[width=0.75\textwidth]{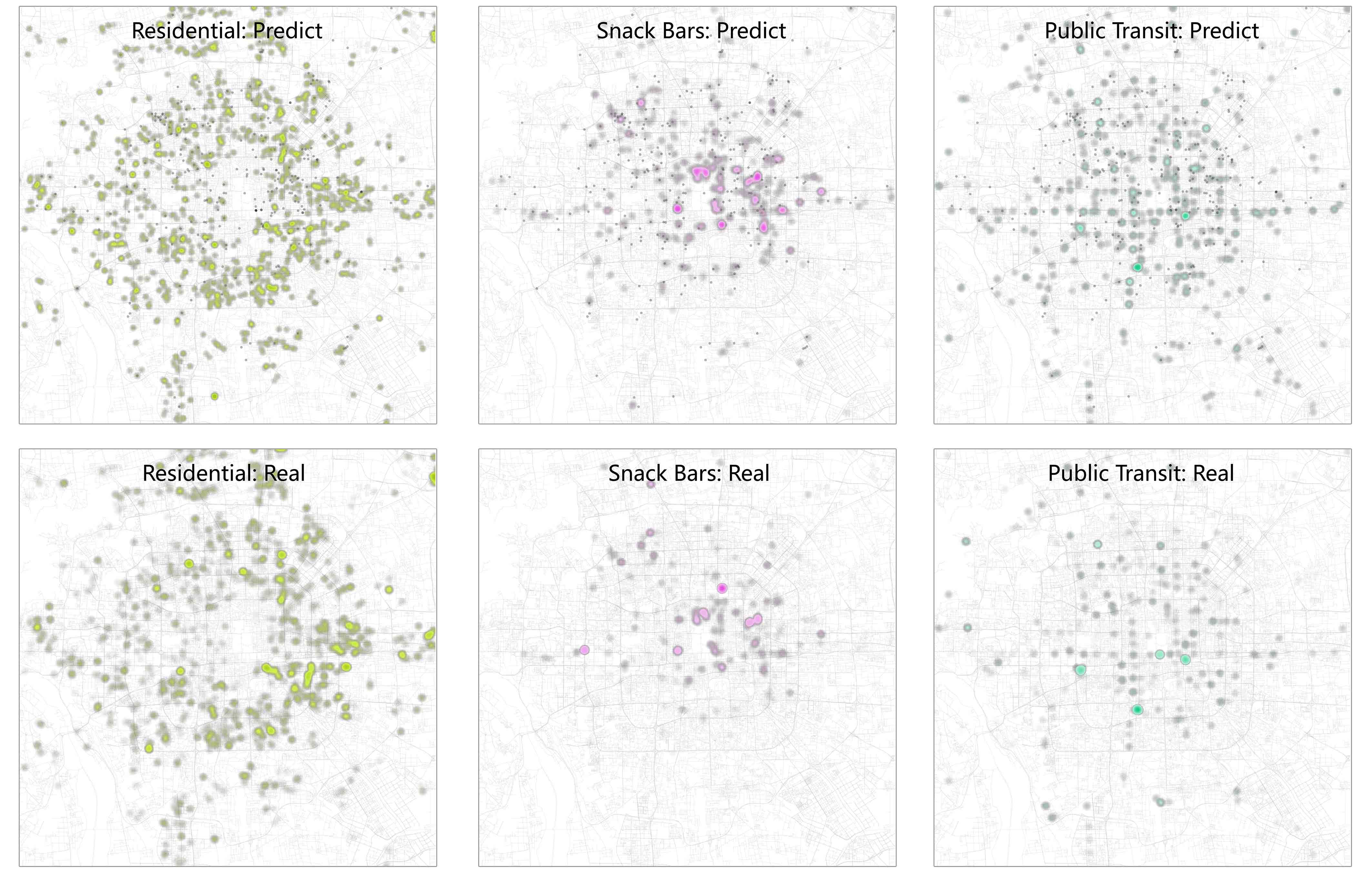}
	\caption{Heatmaps of the predicted check-in patterns and the real check-in patterns for different POI types.}
	\label{fig:predicted_results_types}
\end{figure}

We choose one result of all 50 random experiments as an example to show the predicted spatial patterns in details. The training L1 loss after 2000 epochs is around 0.06 and the absolute error of check-in prediction is about 23, as is shown in Figure \ref{fig:predicted_all_a}. The heatmap of check-in distribution patterns are visualized in Figure \ref{fig:predicted_all_b}. The grey points beneath the predicted pattern are the initialized training points and the color of urban regions indicates the check-in intensity (the redder the higher). Furthermore, the comparison of predicted and real spatial patterns for each POI types are mapped in Figure \ref{fig:predicted_results_types}. It can be seen that our model performs well on all the types, reproducing similar spatial patterns compared with the real ones. Our predicted spatial patterns can reflect the hot spots in urban areas for different functional types.

\section{Discussion}
The experiments conducted in Section \ref{section:experiment} show that our proposed GCN framework is able to reproduce the spatial heterogeneity pattern of intra-urban check-ins based on a small portion of training locations, which demonstrates the feasibility of using graph convolutional neural networks to model irregular spatial patterns. 

\begin{figure}[!htp]
	\subfigure[All POIs (Real)]{\includegraphics[width=0.5\textwidth]{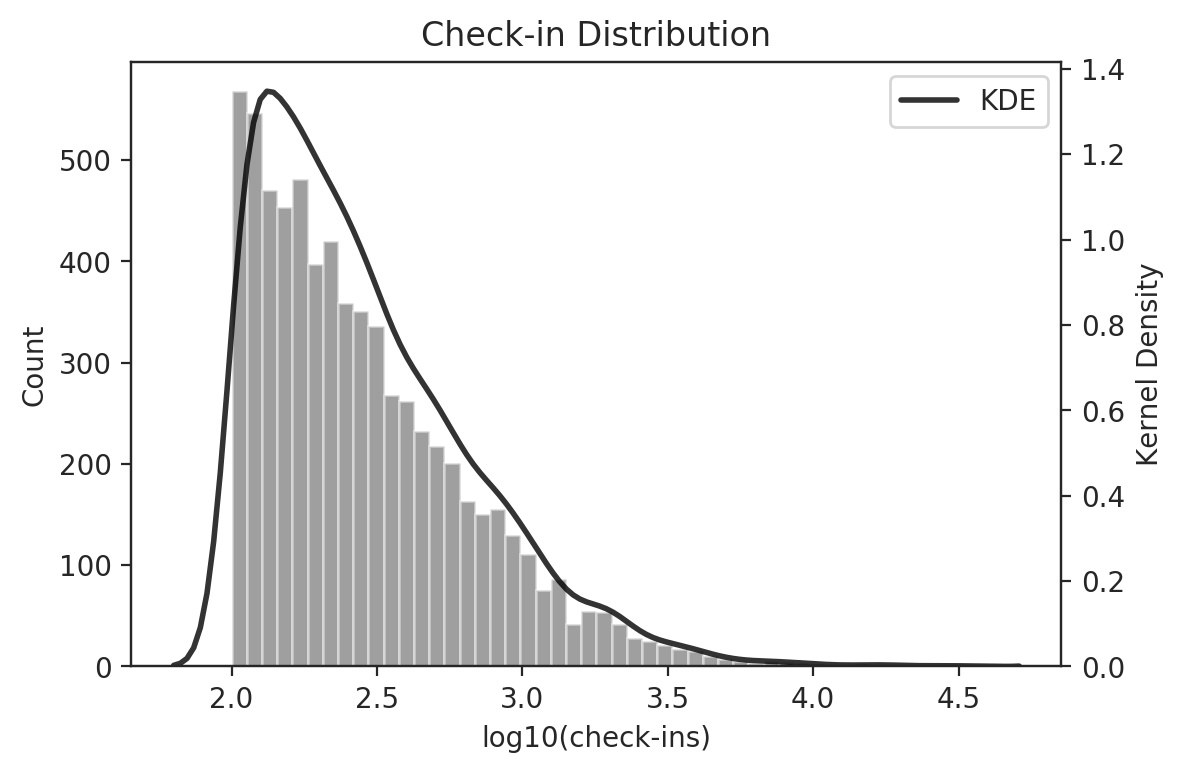}}
	\subfigure[All POIs (Predicted)]{\includegraphics[width=0.5\textwidth]{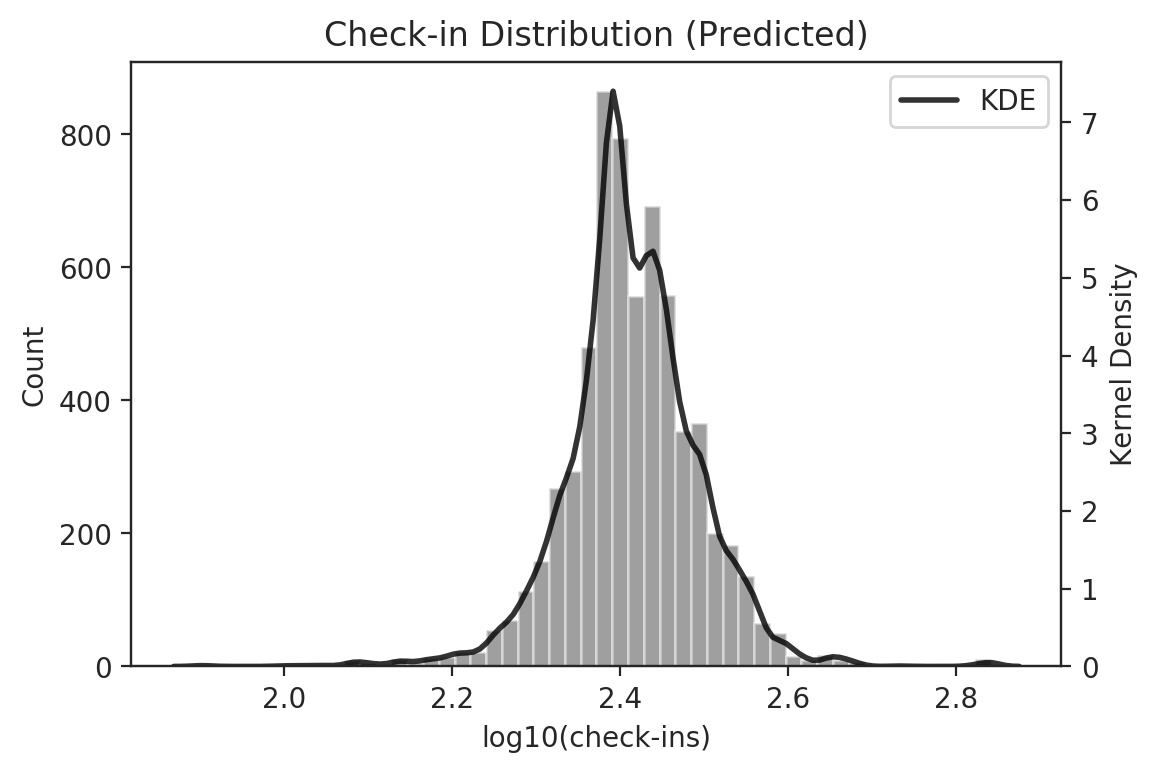}}
	\caption{Statistical distribution of the predicted and real check-in numbers}
	\label{fig:predicted_results_dist}
\end{figure}

\begin{figure}[!htp]
	\includegraphics[width=0.8\textwidth]{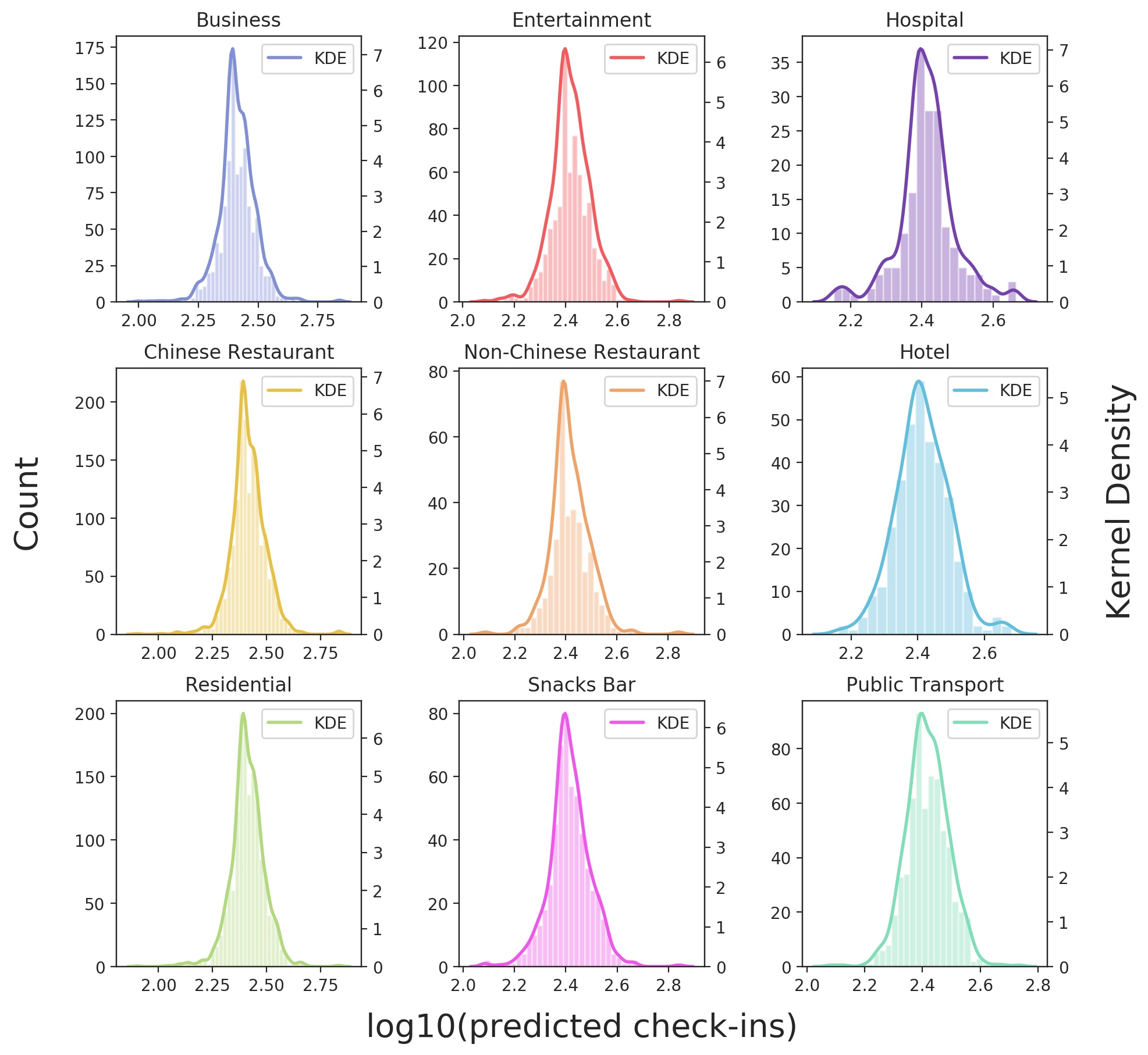}
	\caption{Statistical distribution of the predicted check-in numbers for each POI type.}
	\label{fig:predicted_results_dist_types}
\end{figure}

However, our work also reveals some deficiencies of the GCN-based model that require further endeavors to overcome. Figure \ref{fig:predicted_results_dist} shows the difference of the predicted distribution and the real distribution of the check-in numbers. In the logarithmic x-coordinate, the real check-ins exhibit a strong heavy-tail distribution while the predicted check-ins appear to be a normal distribution, which means that although the model can generate roughly correct spatial patterns of check-ins (Section \ref{sec:predicted_patterns}), the statistical distribution of the neural networks' regression output is hard to approximate non-normal distributions such as power or exponential distributions. Figure \ref{fig:predicted_results_dist_types} further illstrates the statistical distributions of the predicted check-in numbers for different POI types. Compared with Figure \ref{fig:data-b}, we find that it is still difficult to accurately predict the numerical attributes of spatial objects using state-of-the-art GCN-based models even though we have enriched the spatial configurations in the graph.

\section{Conclusions}
In this article, we introduced a generalized model that can capture the spatial pattern in geographical data using graph convolutional networks. By embedding the feature information and the spatial information separately into the graph network, and designing a feature-based localized filter on the graph, our model can learn the spatial dependence among spatial objects and approximate the high-dimensional parameters of spatial patterns according to certain training objectives. 

Based upon that, we proposed a trainable semi-supervised framework using spatial-enriched graph convolutional neural networks to demonstrate the feasibility of our model to be adopted in the analysis of irregular spatial data. The objective is to learn the underlying spatial dependence and predict the unknown features of locations based on their surrounding spatial configurations. The GCN-based framework achieves satisfying results in the prediction of intra-urban POI check-in patterns, and can be modified to be applied to other geographical applications such as spatial interpolation, site selection and event detection.

Important open questions remain: How to evaluate the model's parameters in a way that is both quantitative, interpretable and intuitive for geographical analysis? How to incorporate more understanding of spatial interactions into the graph-based model except for the distance? In addition, this initial work has only focused on the type feature in a single dataset; a promising area is to integrate the features of multi-sourced geo-data such as street networks, remote sensing spectra and other social sensing datasets. An improved version of our model is needed to characterize and explain the intertwined spatial variation pattern in the complex geographic world. We plan to address these questions in on-going works.

\appendix



\bibliography{gcn-arxiv}

\begin{thebibliography}{10}

\bibitem{Anselin1992SPATIAL}
Luc Anselin.
\newblock Spatial data analysis with gis: An introduction to application in the
  social sciences.
\newblock {\em Ncgia Technical Reports}, 1992.

\bibitem{cressie_origins_1990}
Noel Cressie.
\newblock The origins of kriging.
\newblock {\em Mathematical geology}, 22:239--252, 1990.

\bibitem{Defferrard2016Convolutional}
Michaël Defferrard, Xavier Bresson, and Pierre Vandergheynst.
\newblock Convolutional neural networks on graphs with fast localized spectral
  filtering.
\newblock {\em arXiv preprint}, page arXiv:1606.09375, 2016.

\bibitem{Fan1997Spectral}
R.~K.~Chung Fan.
\newblock {\em Spectral graph theory}.
\newblock American Mathematical Society, 1997.

\bibitem{fotheringham2008sage}
A~Stewart Fotheringham and Peter~A. Rogerson.
\newblock {\em The SAGE handbook of spatial analysis}.
\newblock SAGE, 2008.

\bibitem{Glorot2010Understanding}
Xavier Glorot and Yoshua Bengio.
\newblock Understanding the difficulty of training deep feedforward neural
  networks.
\newblock {\em Journal of Machine Learning Research}, 9:249--256, 2010.

\bibitem{Goodchild1992Geographical}
Michael~F. Goodchild.
\newblock Geographical data modeling.
\newblock {\em Computers \& Geosciences}, 18(4):401--408, 1992.

\bibitem{Goodchild2007Towards}
Michael~F. Goodchild, May Yuan, and Thomas~J. Cova.
\newblock Towards a general theory of geographic representation in gis.
\newblock {\em International Journal of Geographical Information Science},
  21(3):239--260, 2007.

\bibitem{Hammond2009Wavelets}
David~K. Hammond, Pierre Vandergheynst, and Rémi Gribonval.
\newblock Wavelets on graphs via spectral graph theory.
\newblock {\em Applied \& Computational Harmonic Analysis}, 30(2):129--150,
  2009.

\bibitem{Henaff2015Deep}
Mikael Henaff, Joan Bruna, and Yann Lecun.
\newblock Deep convolutional networks on graph-structured data.
\newblock {\em arXiv preprint}, page arXiv:1506.05163, 2015.

\bibitem{Kingma2014Adam}
Diederik Kingma and Jimmy Ba.
\newblock Adam: A method for stochastic optimization.
\newblock {\em Computer Science}, 2014.

\bibitem{Kipf2016Semi}
Thomas~N. Kipf and Max Welling.
\newblock Semi-supervised classification with graph convolutional networks.
\newblock {\em arXiv preprint}, page arXiv:1609.02907, 2017.

\bibitem{Lecun2015Deep}
Yann LeCun, Yoshua Bengio, and Geoffrey Hinton.
\newblock Deep learning.
\newblock {\em Nature}, 521(7553):436--444, 2015.

\bibitem{Liu2008Towards}
Yu~Liu, Michael.~F Goodchild, Qinghua Guo, Yuan Tian, and Lun Wu.
\newblock Towards a general field model and its order in gis.
\newblock {\em International Journal of Geographical Information Science},
  22(6):623--643, 2008.

\bibitem{liu2015social}
Yu~Liu, Xi~Liu, Song Gao, Li~Gong, Chaogui Kang, Ye~Zhi, Guanghua Chi, and
  Li~Shi.
\newblock Social sensing: A new approach to understanding our socioeconomic
  environments.
\newblock {\em Annals of the Association of American Geographers},
  105(3):512--530, 2015.

\bibitem{Long2013How}
Ying Long and Xingjian Liu.
\newblock How mixed is beijing, china? a visual exploration of mixed land use.
\newblock {\em Environment \& Planning A}, 45(12):2797--2798, 2013.

\bibitem{Matheron1963Principles}
Georges Matheron.
\newblock Principles of geostatistics.
\newblock {\em Economic Geology}, 58(8):1246--1266, 1963.

\bibitem{Ord1995Local}
J.~K. Ord and Arthur Getis.
\newblock Local spatial autocorrelation statistics: Distributional issues and
  an application.
\newblock {\em Geographical Analysis}, 27(4):286--306, 1995.

\bibitem{DiZhu2018Inferring}
Di~Zhu, Zhou Huang, Li~Shi, Lun Wu, and Yu~Liu.
\newblock Inferring spatial interaction patterns from sequential snapshots of
  spatial distributions.
\newblock {\em International Journal of Geographical Information Science},
  32(4):783--805, 2018.

\end{thebibliography}

\end{document}